\documentclass[conference]{IEEEtran}
\IEEEoverridecommandlockouts

\usepackage{cite}
\usepackage{amsmath,amssymb,amsfonts}
\usepackage{algorithmic}
\usepackage{graphicx}
\usepackage{textcomp}
\usepackage{booktabs}
\usepackage{makecell}
\usepackage{tabularx}
\usepackage{ragged2e}
\usepackage{pgfplots}
\usepackage{hyperref}
\hypersetup{
    hidelinks,
}
\usepackage{xcolor}
\pgfplotsset{compat=1.18}
\def\BibTeX{{\rm B\kern-.05em{\sc i\kern-.025em b}\kern-.08em
    T\kern-.1667em\lower.7ex\hbox{E}\kern-.125emX}}
\begin{document}

\title{
CASE: An Agentic AI Framework for Enhancing Scam Intelligence in Digital Payments
}

\author{
\IEEEauthorblockN{
Nitish Jaipuria\textsuperscript{1},
Lorenzo Gatto\textsuperscript{2},
Zijun Kan\textsuperscript{3},
Shankey Poddar\textsuperscript{4},
Bill Cheung\textsuperscript{5},\\
Diksha Bansal\textsuperscript{6},
Ramanan Balakrishnan\textsuperscript{7},
Aviral Suri\textsuperscript{8},
Jose Estevez\textsuperscript{9}}

\IEEEauthorblockA{\textit{\large Google, Inc}}

\IEEEauthorblockA{
\textit{\small \{
nkjaipuria\textsuperscript{1},
gatto\textsuperscript{2},
zjkan\textsuperscript{3},
shankey\textsuperscript{4},
billcheung\textsuperscript{5},
bdiksha\textsuperscript{6},
ramananb\textsuperscript{7},
aviralsuri\textsuperscript{8},
jmestevez\textsuperscript{9}
\}@google.com}}
}

\maketitle

\begin{abstract}
The proliferation of digital payment platforms has transformed commerce, offering unmatched convenience and accessibility globally. However, this growth has also attracted malicious actors, leading to a corresponding increase in sophisticated social engineering scams. These scams are often initiated and orchestrated on multiple surfaces outside the payment platform, making user and transaction-based signals insufficient for a complete understanding of the scam's methodology and underlying patterns, without which it is very difficult to prevent it in a timely manner. This paper presents CASE (Conversational Agent for Scam Elucidation), a novel Agentic AI framework that addresses this problem by collecting and managing user scam feedback in a safe and scalable manner. A conversational agent is uniquely designed to proactively interview potential victims to elicit intelligence in the form of a detailed conversation. The conversation transcripts are then consumed by another AI system that extracts information and converts it into structured data for downstream usage in automated and manual enforcement mechanisms. Using Google's Gemini family of LLMs, we implemented this framework on Google Pay (GPay) India. By augmenting our existing features with this new intelligence, we have observed a 21\% uplift in the volume of scam enforcements. The architecture and its robust evaluation framework are highly generalizable, offering a blueprint for building similar AI-driven systems to collect and manage scam intelligence in other sensitive domains.
\end{abstract}


\section{Introduction}

The growth of digital payments, exemplified by India's Unified Payments Interface (UPI) \cite{b2}, has attracted malicious actors devising new social engineering scams. A key challenge is that these scams are often orchestrated on external surfaces like messaging apps \cite{b17}, creating a critical intelligence gap. Anti-abuse systems, relying on on-platform signals, lack the context to understand the scam's modus operandi (MO), delaying timely and effective protections.

While empowering users to report experiences is a reliable way to bridge this gap \cite{b15}, traditional static feedback systems (e.g., surveys or free-form text boxes) are insufficient. They fail to capture nuanced details and do not allow for dynamic, clarifying follow-up questions, often yielding ambiguous, non-actionable reports. While manual interviews are effective, they are prohibitively expensive to scale. This necessitates a scalable conversational AI framework that can safely and empathetically conduct these sensitive interviews and process the unstructured conversational data for downstream systems.

In this paper, we present CASE (\textbf{C}onversational \textbf{A}gent for \textbf{S}cam \textbf{E}lucidation), a novel Agentic AI framework designed using the Gemini family of models on Google Pay India. It features two core components: 
\begin{itemize}
\item A \textbf{Conversational Agent}, which leverages Responsible AI principles to safely interview potential victims 
\item An \textbf{Information Extractor Agent}, which processes the unstructured conversation into structured, actionable data. 
\end{itemize}

By augmenting existing features with this intelligence, we demonstrate a 21\% uplift in scam enforcements, all while adhering to a high bar for safety and utility. The remainder of this paper details this framework, its evaluation and its impact, offering a blueprint for similar AI-driven intelligence systems in other payment platforms.

\section{Related Work}

Our work is built upon advancements in four key domains. First, prior AI in scam detection has primarily focused on on-platform signals like account and transaction patterns \cite{b8}\cite{b10}. These models are less effective against social engineering scams where the transaction appears authentic because the core manipulation occurs on external platforms, creating a critical \textit{intelligence gap}. While this narrative intelligence can be gathered via non-scalable and cost prohibitive manual interviews \cite{b14}, our work provides a scalable, automated method for this collection. 

Second, we leverage goal-oriented conversational AI. While LLMs have advanced such systems beyond simple scripted bots \cite{b7}, their application has centered on transactional or informational queries \cite{b7}, not the high-stakes, emotionally sensitive task of conducting investigative, empathetic interviews with potential fraud or scam victims.

Third, to process the resulting transcripts, we utilize modern Information Extraction (IE) techniques. We leverage the paradigm shift enabled by LLMs, which allows for flexible, few-shot extraction from natural language descriptions, forgoing the large, task-specific training datasets required by traditional deep learning approaches \cite{b3}. We apply this to noisy, real-world conversational data, a domain more challenging than typical IE benchmarks. 

Finally, deploying such a system in a high-stakes domain necessitates strict adherence to Responsible AI principles \cite{b13}. Our methodology builds on established best practices by incorporating multi-layered safety architectures, adversarial Red Teaming \cite{b5}, and a hybrid evaluation framework that combines automated tests with expert human review to establish a reliable ground truth \cite{b16} for safety and quality.

\section{Objective}

The primary objective of this paper is to develop and validate an agentic framework, comprising multiple LLMs, for the collection and management of actionable scam intelligence within a large-scale digital payments ecosystem. We specifically focus on the UPI interface in India, using the Google Pay (GPay) India payments platform as our primary implementation case study.

To achieve this, the CASE framework is designed to meet three key system objectives:

\begin{itemize}
\item \textbf{Conversational Agent (Enhance actionable scam intelligence collection):} To develop and deploy a novel conversational agent for proactive information elicitation. This agent must provide a user-friendly and safe interaction that reliably captures nuanced details about social engineering scams, moving far beyond the limitations of traditional, static feedback channels.
\item \textbf{LLM Information Extractor (Enable seamless utilization):} Design and build a powerful backend agent capable of transforming raw, unstructured conversational transcripts into structured, machine-readable data. This ensures that the rich, qualitative insights gathered from user interactions are made programmatically actionable for automated analysis and downstream enforcement systems.
\item \textbf{Establish a Generalizable Framework:} To design the system architecture and evaluation methodologies in a modular and adaptable manner, creating a reusable blueprint for other similar platforms and domains.
\end{itemize}

By fulfilling these technical objectives, we aim to leverage the resulting high-quality, structured intelligence to significantly augment existing fraud detection mechanisms, enabling more precise and scalable enforcement actions against malicious actors.

\section{System Architecture}

This section describes in detail our CASE framework, which is built using Google's Gemini models, and is composed of multiple AI systems designed for scalable scam intelligence collection while adhering to stringent safety and quality standards. It comprises two core components: a user-facing \textit{Conversational Agent} (with an LLM Interviewer and Safety Filter) for real-time data collection and a backend \textit{LLM Information Extractor} for asynchronous data processing.

\begin{figure*}[t!]
\centering
\includegraphics[width=\textwidth]{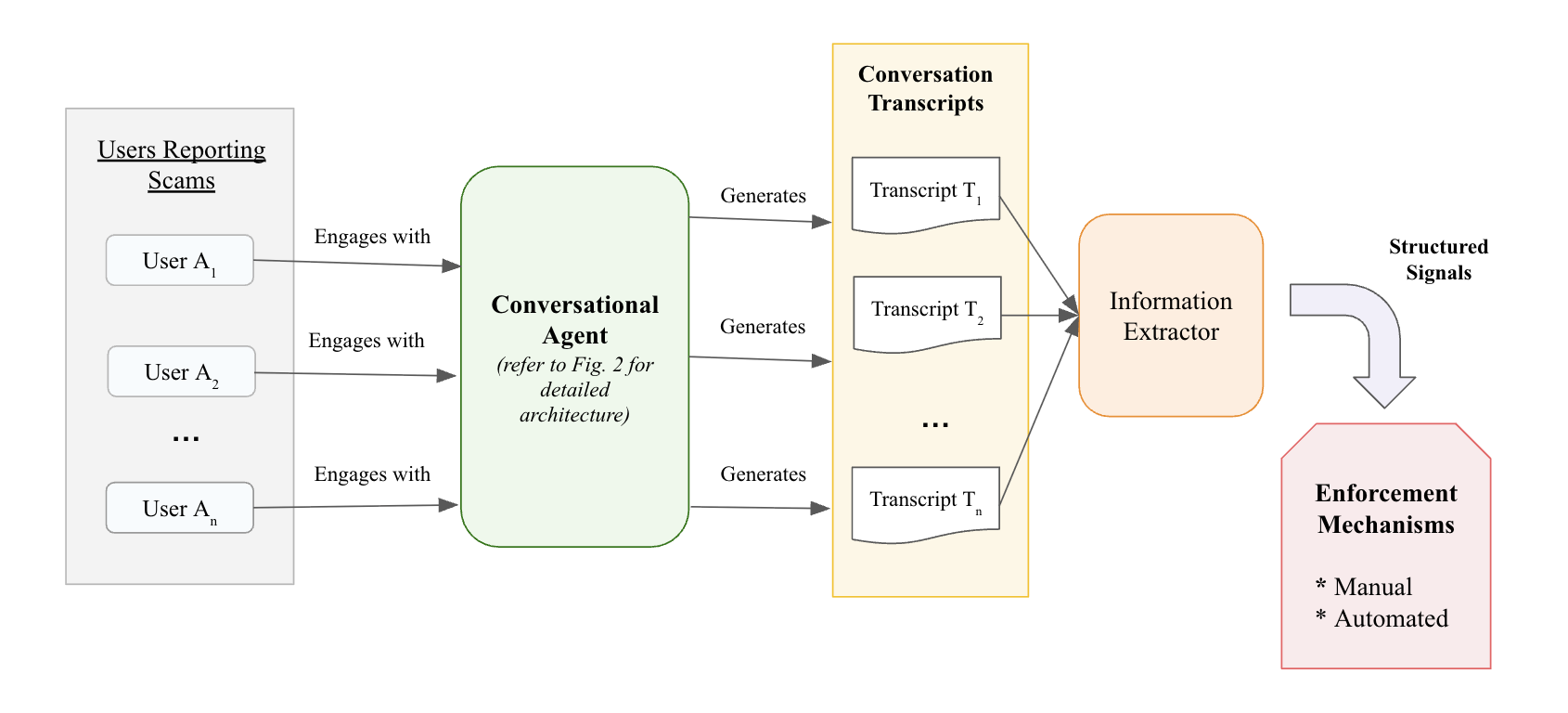}
\caption{Overall System Flow}
\label{fig:fig1}
\end{figure*}

\subsection{End-to-End System Flow}

The system's operational flow (visualized in Figure 1) is integrated into the user's in-app support interface and consists of two phases: 
\begin{itemize}
\item The real-time \textbf{Intelligence Collection} Phase begins when a user's fraud report invokes the LLM service. For each conversational turn, the user's input is processed in parallel by the \textit{Safety Filter LLM} (to assess for policy violations) and the \textit{Generator LLM} (to formulate a response based on conversation history). A decision logic module then evaluates both outputs to determine the final response sent to the user. The dialogue continues until the agent generates a termination token and the full transcript is stored. 
\item The asynchronous \textbf{Data Processing} Phase begins. This is handled as a batch process for reliability and fault tolerance. In this phase, each stored transcript is passed to the Information Extractor Agent, which generates a structured data output based on a predefined schema. This structured intelligence is then written to a dedicated data store, making it available for manual and automated enforcement systems.
\end{itemize}

\subsection{Models Used}
The CASE framework utilizes Google's \textbf{Gemini 2.0 Flash} model \cite{b1} for all three core components: the real-time Interview Agent, the LLM Safety Filter and the asynchronous Information Extractor Agent. This high-efficiency model was chosen for production as its balance of speed, capability and cost is ideal for large-scale deployment. During the research and development phase, more powerful models like \textbf{Gemini 2.0 Pro} \cite{b1} were also used to establish performance benchmarks and explore upper bounds of capability.

\begin{figure*}[t!]
\centering
\includegraphics[width=\textwidth]{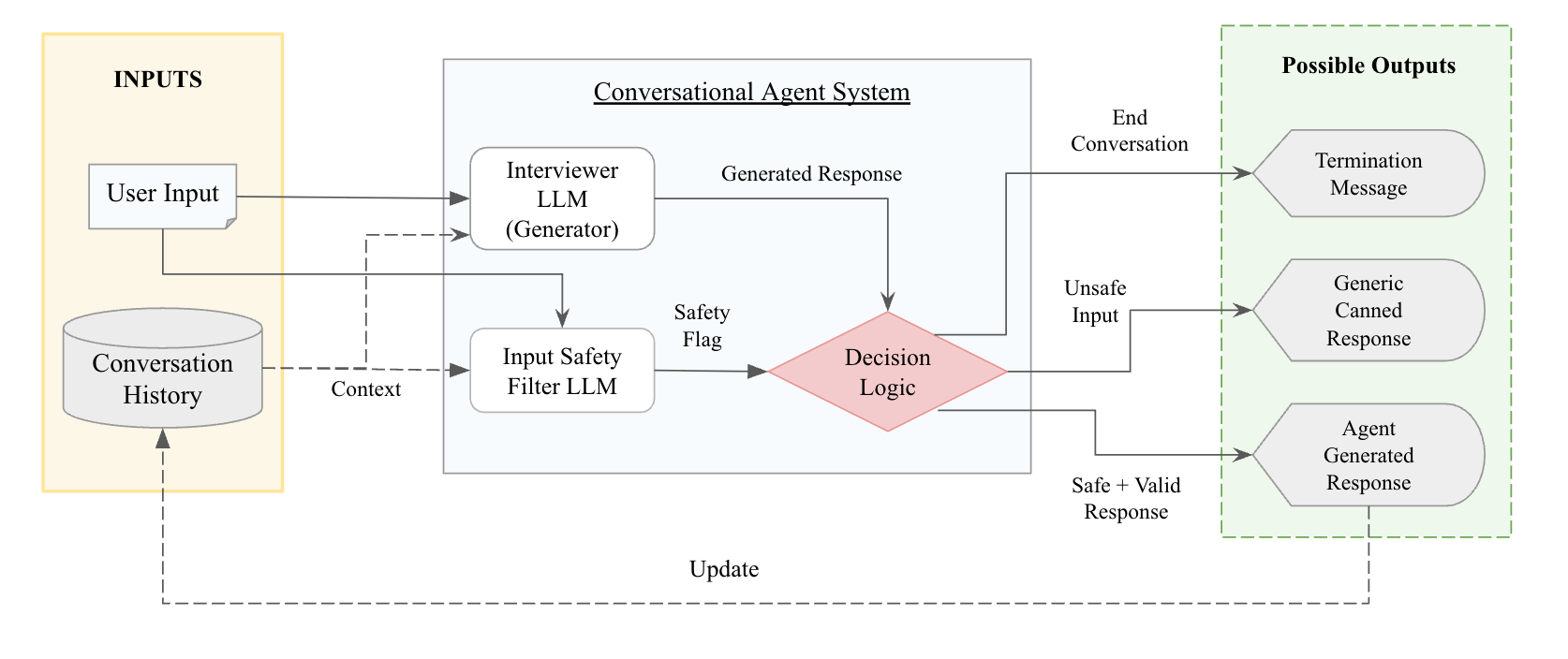}
\caption{Conversation Agent Flow}
\label{fig:fig2}
\end{figure*}

\subsection{Conversational Agent: Generator LLM}
The Conversational Agent's primary objective is to conduct a reliable, useful and safe user interview to understand a potentials scam's complete modus operandi. Unlike typical chatbots that \textit{answer} questions, its function is to \textit{ask} dynamic, investigative questions. As detailed in Figure 2, the agent is powered by the Gemini Flash 2.0 model \cite{b1}. Its behavior is guided by a sophisticated prompt architecture, chosen over fine-tuning for its flexibility and iteration speed. The prompt's core components are: 
\begin{itemize}
\item \textit{Role Definition}: instructing the LLM to adopt the persona of a specialist fraud analyst grounded with domain-specific examples.
\item \textit{Interaction Guidelines}: a strict set of rules such as: never promising a refund, avoiding financial advice, staying on topic and immediately respecting a user's wish to stop.
\item \textit{Success Criteria}: which defines a successful interview as one that captures key facets of the scam (e.g., initial contact surface or the hook used to build trust etc.) 
\item \textit{Privacy by Design}: ensuring prompts are stateless and contain no PII, as any user or transaction data is not passed to the LLM.
\end{itemize}

\subsection{Conversational Agent: Safety Filter LLM}
Given the sensitive nature of financial scam discussions, the CASE framework incorporates a robust, multi-layered safety architecture. This consists of three layers working in concert: 
\begin{itemize}
\item \textbf{Base Model Safeguards} integrated into the Gemini model \cite{b1} to prevent harmful content generation 
\item \textbf{Input Safety Filter}: a dedicated LLM-based classifier that runs in parallel with the main generator to assess all user inputs for policy-violating content and 
\item \textbf{Guided Prompt Architecture} with strict negative constraints to prevent contextually inappropriate outputs and to gracefully conclude off-topic interactions.
\end{itemize}

\begin{table*}[t!]
\caption{A subset of the data schema populated by the Information Extractor.}
\label{tab:extractor_schema}
\centering
\begin{tabularx}{\textwidth}{@{} ll >{\RaggedRight}X >{\RaggedRight}X @{}}
\toprule
\textbf{Feature Name} & \textbf{Requirement} & \textbf{Description} & \textbf{Possible Values} \\
\midrule
\texttt{is\_user\_scammed} & Mandatory & A boolean flag indicating if the agent determined the user was a victim of a scam. & \texttt{True}, \texttt{False} \\
\addlinespace
\texttt{possible\_scam\_mo} & Mandatory & The classified Modus Operandi (MO) of the scam. Choose \texttt{NOT\_SCAM} only when \texttt{is\_user\_scammed} = False. & \texttt{NOT\_SCAM}, \texttt{FAKE\_LOAN}, \texttt{FAKE\_JOBS}, \texttt{FAKE\_ADS}, ... \\
\addlinespace
\texttt{scam\_origin\_surface} & Optional & The application or surface where the user first came into contact with the scammer. & \texttt{<List of popular social media and messaging apps>}, \texttt{OTHERS}, \texttt{NONE}, ... \\
\addlinespace
\texttt{conversation\_summary} & Mandatory & A concise, LLM-generated summary of the entire interview. & Free-form text within word limits. \\
\bottomrule
\end{tabularx}
\end{table*}

\subsection{LLM Information Extractor}
The Information Extractor Agent converts unstructured conversational transcripts into a structured, machine-readable format, making the qualitative insights programmatically actionable. Implemented using the \textit{Gemini Flash 2.0} model, its behavior is guided by a sophisticated prompt architecture, chosen for its flexibility over fine-tuning or traditional NLP pipelines. This prompt combines role-based instructions with a schema-guided in-context learning (ICL) strategy, using diverse, high-quality examples from a manually annotated \textbf{golden dataset} \cite{b9} to ensure high-fidelity output. The agent's task is to populate a predefined, adaptable data schema, a subset of which is shown in Table I, with a combination of mandatory and optional features, outputting them as a JSON object for ingestion by downstream risk systems. Figure 3 illustrates a sample input-output pair, through an example.

\begin{figure*}[t!]
\centering
\includegraphics[width=\textwidth]{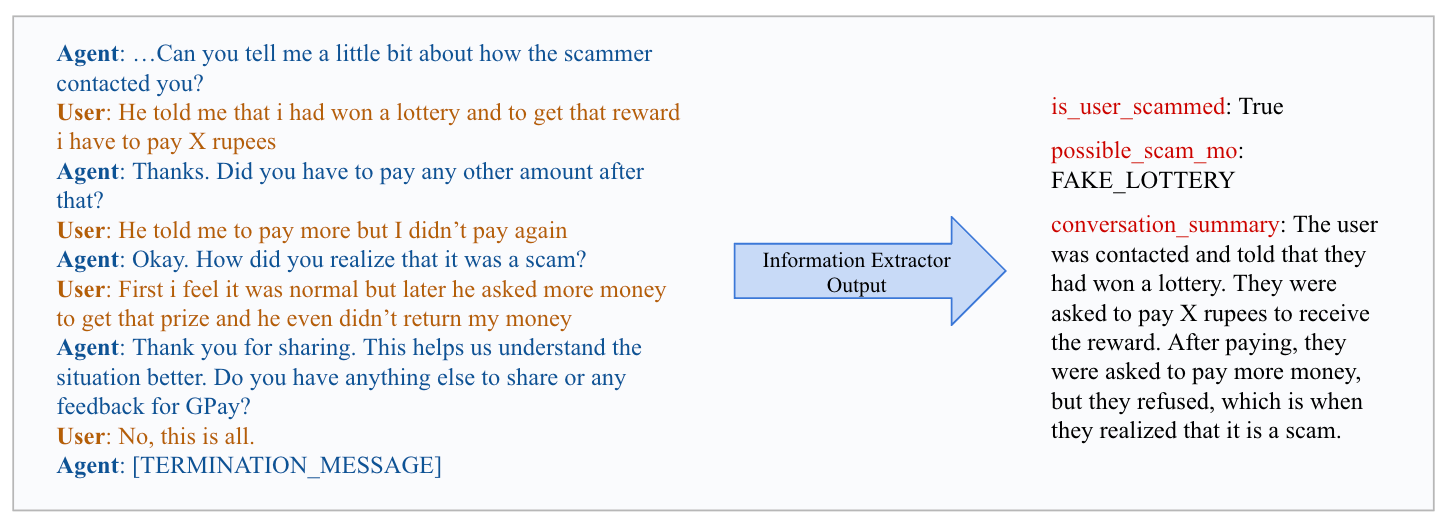}
\caption{Conversation Transcript to Information Extractor Structured Output Examples}
\label{fig:fig3}
\end{figure*}

\subsection{Scam Intelligence Utilization}
The structured data generated by the Information Extractor is actively integrated into the anti-abuse ecosystem in two primary ways. First, it is made available to human reviewers for deep-dive investigations into complex cases, helping analysts understand new scam MOs and validate high-value takedowns. Second, at an aggregated level, the structured data serves as a powerful new set of signals for machine learning risk models, allowing the intelligence gathered from user experiences to automatically detect and prevent new scam patterns at scale. This dual-use approach, combining immediate, human-led review with long-term, automated learning, ensures every piece of user-provided intelligence contributes to platform safety.

\section{Evaluation Methodology}
Given the system's sensitive, user-facing nature, our evaluation methodology ensures safety, reliability, and effectiveness across the entire product lifecycle. This framework rests on the pillars of Safety (preventing harm), Quality \& Utility (achieving system goals), with distinct plans for the Conversational Agent and Information Extractor, and is designed to transition from human raters to automated evaluators for scalable, long-term monitoring.

\subsection{Expert Human Raters}
Given the sensitive nature of user interactions and the complexity of assessing subjective qualities, human evaluations are a foundational component of our testing framework, critical for establishing ground truth and validating performance in scenarios challenging to automate \cite{b16}. 

To ensure a qualified assessment, we utilize a specialized group of domain expert human raters. This group consists of professionals trained to review accounts for payments scam signals and take appropriate enforcement actions. To qualify them for this specific task, they were additionally trained on a set of curated guidelines and rubrics for evaluating the Conversational Agent's safety and quality. These experts were instrumental during both pre-launch development and post-launch monitoring, providing qualitative insights, conducting adversarial tests (Red Teaming) \cite{b5}, and validating system behavior against our standards \cite{b11}.

\subsection{Conversational Agent: Safety}
Our safety evaluation is built upon rigorously defined policies, which are categorized into two tiers. The first tier targets \textbf{high-severity} violations like hate speech, dangerous content etc. The second, more nuanced tier evaluates \textbf{contextually sensitive} content which are typically not considered harmful but are contextually inappropriate for our system, such as providing financial advice. 

We used two primary methods to test against these policies: \textit{Structured Evals}, which used large, curated datasets of adversarial prompts sourced from both central safety teams and handcrafted, context-specific examples; and \textit{Red Teaming}, where human experts simulated real-world bad actors to discover unknown vulnerabilities \cite{b5}.

\subsection{Conversational Agent: Quality and Utility}
Beyond safety, we evaluated the agent's core function by assessing the \textbf{Quality} of the conversational experience and the \textbf{Utility} of the information gathered. Quality metrics include \textit{Topic Adherence} which is the agent's ability to remain focused and gracefully handle off-topic inputs and \textit{User Respect} or maintaining a supportive tone and immediately concluding if the user wishes to stop. 

The Utility metric, Information Elicitation, measured the agent's success in capturing the scam's core modus operandi aka \textit{possible\_scam\_mo}. In the pre-production stage, these subjective metrics were scored by our trained human evaluators, who assessed sample conversations against detailed guidelines for consistent scoring.

\subsection{LLM Information Extractor: Accuracy}
In contrast to the subjective agent evaluation, the Information Extractor's assessment is more quantitative and is framed as a series of classification problems. To establish the necessary ground truth, we utilized a hand-crafted \textbf{golden dataset} of manually curated conversation transcripts and the corresponding structured extraction data. The agent's accuracy was then measured, focusing on the two features most critical for enforcement: 
\begin{itemize}
\item \textbf{Scam Detection}: treated as a binary classification task corresponding to the \textit{is\_user\_scammed} field
\item \textbf{Modus Operandi} Classification: a multiclass classification for the \textit{possible\_scam\_mo} field.
\end{itemize}

While this methodology can be extended, the numerous optional fields were not formally scored in this initial implementation.

\subsection{Post-Production Monitoring}
Our evaluation framework extends into a continuous post-production monitoring process as outlined in Figure 4. Following a gradual rollout where a high proportion of conversations were manually reviewed to establish a real-world baseline, the methodology transitions to a scalable, steady-state hybrid approach. In this approach, an automated evaluator aka \textit{autorater} \cite{b12} assesses all the conversations, while a statistically significant sample is concurrently routed to human evaluators. The results from both are continuously compared to validate and calibrate the autorater's performance and to identify novel issues, ensuring both long-term quality control and operational scalability.

\begin{figure*}[t!]
\centering
\includegraphics[width=\textwidth]{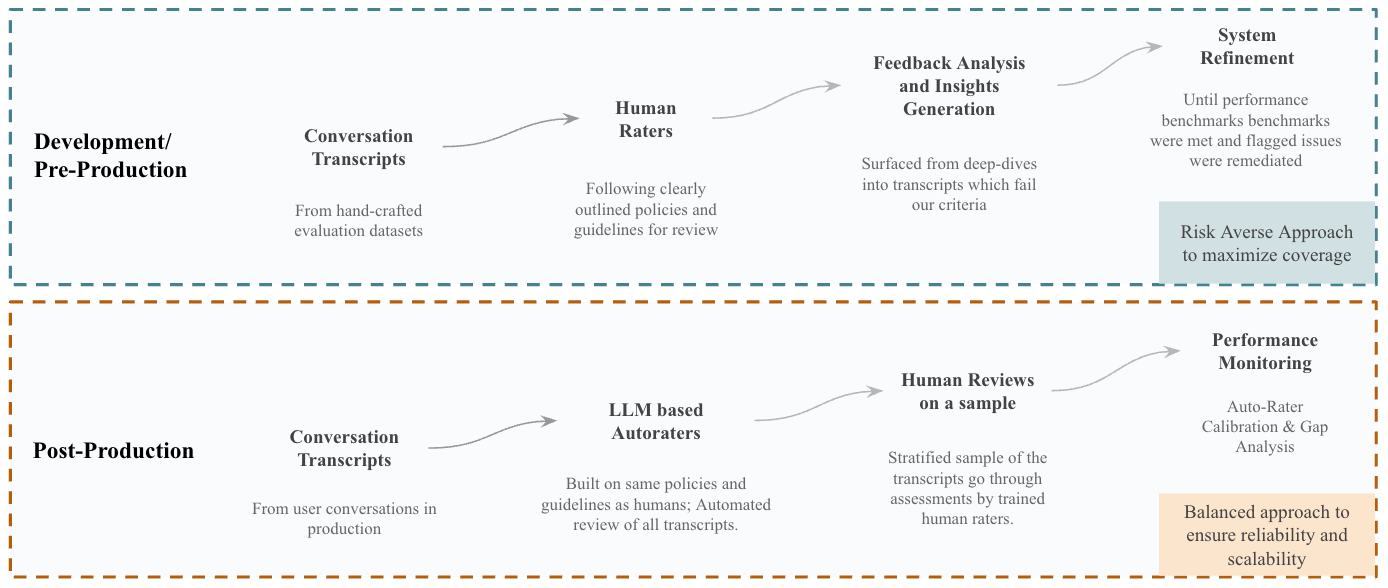}
\caption{Evaluation Process Flow Comparison Pre and Post Production.}
\label{fig:fig4}
\end{figure*}

\section{Results and Impact}
The CASE framework has been deployed within the Google Pay India ecosystem. The post-launch results presented here were collected during a partial production roll-out and, wherever applicable, have been extrapolated to project the full-scale impact. The system underwent continuous iterative development and the pre-production metrics reported reflect the state of the system immediately prior to launch. The results validate the system's effectiveness across its core performance, intelligence gathering and downstream enforcement.

\subsection{User Engagement}
The agent enabled the collection of a new, rich intelligence source at scale with strong user engagement. We analyzed user engagement by visualizing the distribution of agent-generated questions answered by users per conversation as shown in Figure 5. While the data shows a natural drop-off as conversation length increases, it also reveals a significant cohort of highly engaged users. The data confirms the agent's effectiveness at maintaining a productive dialogue, with over 45\% of the conversation sessions involving users answering three or more follow-up questions, exceeding our target for what constitutes a meaningful, in-depth interview.

\subsection{Agent Performance and Enforcement Impact}
The agent's performance was validated across all three key areas. For Safety, pre-launch structured evaluations achieved 99.9\% compliance on \textit{high severity} violation policies and 99.2\% compliance on \textit{contextually sensitive topic} policies, exceeding standard benchmarks \cite{b4}\cite{b11}. This was validated in production, on a sample size of approximately 3600 samples, which recorded \textbf{no} (i.e. 100\% compliance on) high severity violations and \textbf{99.5\%} compliance on contextually sensitive topic violations. For Quality and Utility, human evaluations confirmed the agent performed strongly against our quality metrics, with detailed results in Table II. Finally, the Information Extractor, verified against a sample of 3000 live production conversations, achieved an \textbf{83.8\%} accuracy on the \textit{is\_user\_scammed} binary classification task and \textbf{75.1\%} accuracy on the multiclass \textit{possible\_scam\_mo} classification task, compared to human reviewer labels.

The incremental and high quality intelligence generated by the framework have a direct, measurable downstream enforcement impact. These new signals when augmented with our existing enforcement mechanisms are projected to increase the overall \textbf{scam detection recall} by \textbf{21\%}.

The CASE framework continues to be enhanced in production, with ongoing analysis of live user conversations providing new insights for further improvements.

\begin{table}[t!]
\caption{Quality and Utility Metrics}
\label{tab:quality_metrics}
\centering
\begin{tabular}{lccc}
\toprule
& \multicolumn{2}{c}{\textbf{Quality}} & \textbf{Utility} \\
\cmidrule(lr){2-3}
& \makecell{Topic \\ Adherence} & \makecell{User \\ Respect} & \makecell{Scam MO \\ Elicitation} \\
\midrule
Pre-Prod & 99.9\% & 99.8\% & - \\
Post-Prod & 99.9\% & 99.9\% & 75.1\% \\
\bottomrule
\end{tabular}
\end{table}

\begin{figure}[t!]
\centering
\includegraphics[width=\columnwidth]{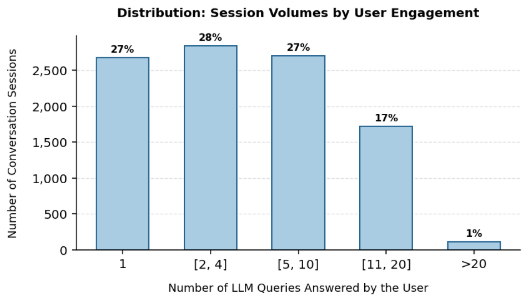}
\caption{Distribution of user sessions based on the number of LLM queries responded to in each session. Percentages indicate the proportion of total sessions for each bucket.}
\label{fig:fig5}
\end{figure}

\section{discussion}
Scams are a significant and growing global threat, creating a need for innovative, scalable AI solutions to combat them early. The deployment and impact of our agentic AI framework demonstrates the potential of this approach in the Trust \& Safety domain. In this section, we interpret our findings, discuss their generalizability and acknowledge its limitations.

\subsection{Key Insights and Interpretation}
Our primary insight is that augmenting traditional, transaction-based signals with structured intelligence from user narratives is a highly effective strategy for combating social engineering scams, which often lack clear on-platform signals. This newly generated intelligence can be utilized as a strong signal in advanced scam and fraud-fighting frameworks, such as those described in \cite{b6}\cite{b18}.

The dual Conversational Agent-Information Extractor pattern proved to be a robust and scalable architecture; separating the real-time, user-facing task from the asynchronous, backend data processing, created a system that is both responsive and efficient. Furthermore, this work underscores that for a sensitive, user-facing application, a comprehensive, multi-layered safety and quality evaluation framework is not merely a preliminary step but a core, necessary component for building trust and ensuring a safe user experience.

\subsection{Generalizability of the CASE Framework}
While the immediate focus of this research is on scam detection within Google Pay India, the architectural design of the proposed LLM-based system supports broad generalization and scalability. This framework is readily adaptable for deployment across various Google Pay and other similar payments platforms where peer-to-peer (P2P) payment functionalities are operational. Preliminary experimentation is already planned for other international markets, requiring small to medium internal adjustments to accommodate regional nuances. Furthermore, the utility of this system extends beyond the Google Pay ecosystem. With appropriate modifications, the core LLM-driven interaction and interpretation engine can serve as a robust instrument for engaging with any abuse-related conversational data, facilitating direct interactions with automated agents. The inherent flexibility of the LLM system allows it to analyze and address a wide spectrum of abusive behaviors. This adaptability is largely facilitated by the strategic customization of prompts and providing domain specific examples, which ground the agent in a new context.

Crucially, this LLM-based classification system incorporates multiple layers of integrated guardrails. These guardrails not only ensure its safe and effective deployment for diverse payment-related use cases but also significantly broaden its applicability. The system's substantial potential for addressing a multitude of other critical use cases, including but not limited to, online harassment, hate speech, misinformation, and various forms of harmful content, positions it as a highly effective conversational agent for a wide array of customer interaction scenarios, thereby extending its utility catering to the evolving Trust and Safety challenges of the present and foreseeable future.

\subsection{Limitations}
We acknowledge several limitations which present opportunities for future research. Firstly, the framework was evaluated for English-only conversations; while the underlying Gemini model has multilingual capabilities \cite{b1}, leveraging this for other languages might require careful prompt design and further testing. Secondly, the system's behavior is governed by sophisticated prompt engineering, which, while flexible, may be less robust than a fully fine-tuned model for handling specific conversational nuances. Finally, our evaluation framework is strategically dependent on human reviews. While a hybrid auto-rater system is used for scaling, maintaining a cohort of trained human evaluators for continuous validation is an essential, long-term component to mitigate the risk of auto-rater performance degradation.

\section{Conclusion}
In this paper, we addressed the critical challenge of insufficient signals for fighting against social engineering scams, by designing and deploying CASE, a novel dual-agent AI framework that automates interviewing potential victims and structuring their narratives into actionable data. Our real-world deployment on Google Pay India demonstrated that this system can safely and effectively engage with users to produce high-volume structured intelligence, leading to direct improvements in enforcement recall and velocity. While acknowledging limitations, such as its initial English-only implementation and evaluation scaling, this research validates the Conversational Agent-Information Extractor pattern as a generalizable blueprint for other Trust \& Safety domains and confirms the strategic value of using agentic AI to augment traditional anti-abuse systems against narrative-driven harms.

\section{Future work}
Future work will focus on three key areas: First, we plan to incorporate \textbf{multimodal inputs}, allowing users to provide evidence via audio recordings or screenshots to create a richer data collection experience. Second, we will continue the development of our \textbf{auto-raters} for more robust steady-state monitoring and explore granting the framework greater autonomy for direct enforcement based on high-confidence outputs. Finally, looking beyond a single platform, we envision the CASE framework as a tool for \textbf{ecosystem-level collaboration}, where anonymized, structured insights on new scam MOs could be shared with external agencies or a consortium of financial institutions for a more coordinated defense against online social engineering scams.

\end{document}